\documentclass[11pt,a4paper]{article}
\usepackage[hyperref]{emnlp-ijcnlp-2019}
\usepackage{times}
\usepackage{latexsym}
\usepackage{graphicx}
\usepackage{amssymb}
\usepackage{amsmath}
\usepackage{booktabs}
\usepackage{float}
\usepackage{framed}
\usepackage{cleveref}
\usepackage{verbatim} 
\usepackage{enumitem}

\aclfinalcopy 

\newcommand{\tabref}[1]{Table \ref{#1}}
\newcommand{\figref}[1]{Figure \ref{#1}}
\renewcommand{\eqref}[1]{Eq.~\ref{#1}}
\newcommand{\appref}[1]{Appendix \ref{#1}}

\definecolor{ao}{rgb}{0.0, 0.5, 0.0}

\crefformat{section}{\S#2#1#3} 
\crefformat{subsection}{\S#2#1#3}
\crefformat{subsubsection}{\S#2#1#3}

\title{Towards Debiasing Fact Verification Models}

\author{Tal Schuster*$^{,1},$ ~~ Darsh J Shah*$^{,1},$ ~~ Yun Jie Serene Yeo$^2,$ \\
\textbf{Daniel Filizzola$^1,$ ~~ Enrico Santus$^1,$ ~~ Regina Barzilay$^1$}\\
  \mbox{}\\
$^1$Computer Science and Artificial Intelligence Lab, MIT \\
$^2$DSO National Laboratories, Singapore \\
\small{\texttt{\{tals, darsh, yeoyjs, danifili, esantus, regina\}@csail.mit.edu}}
}

\date{}

\begin{document}

\maketitle
\let\svthefootnote\thefootnote
\let\thefootnote\relax\footnote{Asterisk (\textbf{*}) denotes equal contribution.}
\addtocounter{footnote}{-1}
\let\thefootnote\svthefootnote
\begin{abstract}
Fact verification requires validating a claim in the context of evidence. We show, however, that in the popular FEVER dataset this might not necessarily be the case. Claim-only classifiers perform competitively with top evidence-aware models. In this paper, we investigate the cause of this phenomenon, identifying strong cues for predicting labels solely based on the claim, without considering any evidence. We create an evaluation set that avoids those idiosyncrasies. The performance of FEVER-trained models significantly drops when evaluated on this test set. Therefore, we introduce a regularization method which alleviates the effect of bias in the training data, obtaining improvements on the newly created test set. This work is a step towards a more sound evaluation of reasoning capabilities in fact verification models.\footnote{Data and code: \url{https://github.com/TalSchuster/FeverSymmetric}}

\end{abstract}
\section{Introduction}

Creating quality datasets is essential for expanding NLP functionalities to new tasks. Today, such datasets are often constructed using crowdsourcing mechanisms. Prior research has demonstrated that 
artifacts of this data collection method often introduce idiosyncratic biases that impact performance in unexpected ways~\citep{poliak2018hypothesis,gururangan2018annotation}. In this paper, we explore this issue using the FEVER dataset, designed for fact verification ~\cite{thorne2018fever}.

\begin{figure}[t]
\small
\begin{center}
  \includegraphics[width=0.4\textwidth]{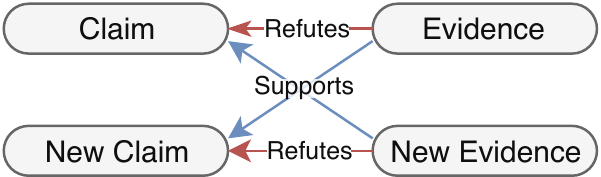}
\end{center}
\begin{framed}
\small
        \vspace*{-0.5\baselineskip}
        \begin{center}\textbf{(A)} \underline{\textsc{Original} pair from the FEVER dataset}  \end{center} 
        \textbf{Claim:} \newline Stanley Williams stayed in Cuba his whole life.
        \newline
        \textbf{Evidence:}\newline  Stanley [...] was part of the West Side Crips, a street gang which has its roots in South Central Los Angeles.

\end{framed}
 \vspace*{-\baselineskip}
\begin{framed}
\small
   \vspace*{-0.5\baselineskip}
        \begin{center}\textbf{(B)} \underline{Manually \textsc{Generated} pair} \end{center}
        \textbf{Claim:} \newline Stanley Williams moved from Cuba to California when he was 15 years old.
        \newline 
        \textbf{Evidence:}\newline Stanley [...] was born in Havana and didn't leave the country until he died. 
\end{framed}
\caption{
An illustration of a \textsc{Refutes} claim-evidence pair from the FEVER dataset (A) that is used to generate a new pair (B). From the combination of the \textsc{Original} and manually \textsc{Generated} pairs, we obtain a total of four pairs creating symmetry.
}
\label{fig:intro}
\vspace*{-\baselineskip}
\end{figure}

The task of fact verification involves assessing claim validity in the context of evidence, which can either support, refute or contain not enough information. \figref{fig:intro}(A) shows an example of a FEVER claim and evidence. While validity of some claims may be asserted in isolation (e.g.\ through common sense knowledge), contextual verification is key for a fact-checking task~\citep{alhindi-etal-2018-evidence}. 
Datasets should ideally evaluate this ability. 
To assess whether this is the case for FEVER, 
we train a claim-only BERT \cite{devlin2018bert} model that classifies each claim on its own, without associated evidence. The resulting system achieves 61.7\%, far above the majority baseline (33.3\%).

Our analysis of the data demonstrates that this unexpectedly high performance is due to idiosyncrasies of the dataset construction. For instance, in \cref{sec:bias_analysis} we show that the presence of negation phrasing highly correlates with the \textsc{Refutes} label, independently of provided evidence.

To address this concern, we propose a mechanism for avoiding bias in the test set construction. We create a \textsc{Symmetric Test Set} where, for each claim-evidence pair, we manually generate a synthetic pair that holds the same relation (e.g.\ \textsc{Supports} or \textsc{Refutes}) but expressing a different, contrary, fact. In addition, we ensure that in the new pair, each sentence satisfies the inverse relation with the original pair's sentence. This process is illustrated in \figref{fig:intro}, where an original \textsc{Refutes} pair is extended with a synthetic \textsc{Refutes} pair. The new evidence is constrained to support the original claim, and the new claim is supported by the original evidence. In this way, we arrive at three new pairs that complete the symmetry.

Determining veracity with the claim alone in this setting would be equivalent to a random guess. Unsurprisingly, the performance of FEVER-trained models drop significantly on this test set, despite having complete vocabulary overlap with the original dataset.  For instance, the leading evidence-aware system in the FEVER Shared Task, the NSMN classifier by \citet{nie2018combining}\footnote{\url{https://github.com/easonnie/combine-FEVER-NSMN}}, achieves only 58.7\% accuracy on the symmetric test set compared to 81.8\% on the original dataset.

While this new test set highlights the aforementioned problem, other studies have shown that FEVER is not the only biased dataset \citep{poliak2018hypothesis,gururangan2018annotation}. A potential solution which may be applied also in other tasks is therefore to develop an algorithm that alleviates such bias in the training data. We introduce a new regularization procedure to downweigh the give-away phrases that cause the bias.

\noindent The contributions of this paper are threefold:
\begin{itemize} [leftmargin=0cm,itemindent=.3cm,labelwidth=\itemindent,labelsep=0cm,align=left]
\setlength\itemsep{-0.1em}
\item We show that inherent bias in FEVER dataset interferes with context-based fact-checking. 
\item We introduce a method for constructing an evaluation set that explicitly tests a  model's ability to validate claims in context.
\item  We propose a new regularization mechanism that improves generalization in the presence of the aforementioned bias.
\end{itemize}

\section{Motivation and Analysis}
\label{sec:bias_analysis}

In this section, we quantify the observed bias and explore the factors causing it. 

\paragraph{Claim-only Classification}

\begin{table}[t]
\centering
\resizebox{\columnwidth}{!}{%
\begin{tabular}{lcc|cc}
\toprule
\textbf{}          & \multicolumn{2}{c|}{\textbf{Train}} & \multicolumn{2}{c}{\textbf{Development}} \\ 
\textbf{Bigram}    & \textbf{LMI$ \cdot 10^{-6}$}   & \textbf{$p(l |w)$}  & \textbf{LMI$\cdot10^{-6}$}     & \textbf{$p(l |w)$}  \\ 
\midrule
did not            & 1478           & 0.83      & 1038             & 0.90       \\
yet to             & 721            & 0.90       & 743              & 0.96      \\
does not           & 680            & 0.78       & 243              & 0.68      \\
refused to         & 638            & 0.87      & 679              & 0.97       \\
failed to          & 613            & 0.88      & 220              & 0.96       \\
only ever          & 526            & 0.86       & 350              & 0.82      \\
incapable being    & 511            & 0.89       & 732              & 0.96      \\
to be              & 438            & 0.50       & 454              & 0.65      \\
unable to          & 369            & 0.88       & 346              & 0.95      \\
not have           & 352            & 0.78       & 211              & 0.92      \\
\bottomrule             
\end{tabular}%
}
\caption{Top 10 LMI-ranked bigrams in the train set of FEVER for \textsc{Refutes} with its $p(l|w)$. The corresponding figures for the development set are also provided. Statistics for other labels are in \appref{appx:top_bigrams}.}
\label{tbl:top-20-lmi-refutes}
\end{table}

Claim-only aware classifiers can significantly outperform all baselines described by \citet{thorne2018fever}.\footnote{We evaluate on the development set as the test set is hidden. Hyper-parameter fine-tuning is performed on a 20\% split of the training set, which is finally joined to the remaining 80\% for training the best setting. See \appref{appx:fever_split}.} BERT, for instance, attains an accuracy of 61.7\%, which is just 8\% behind NSMN.
We hypothesize that these results are due to two factors: (1) idiosyncrasies distorting performance and (2) word embeddings revealing world knowledge.

\paragraph{Idiosyncrasies Distorting Performance} We investigate the correlation between phrases in the claims and the labels. In particular, we look at the n-gram distribution in the training set. 
We use Local Mutual Information (LMI)~\citep{evert2005statistics} to capture high frequency n-grams that are highly correlated with a particular label, as opposed to $p(l |w)$ that is biased towards low frequency n-grams. LMI between $w$ and $l$ is defined as follows: 
\begin{equation}\label{eq:lmi}
    LMI(w, l)  = p(w,l) \cdot \log \bigg(\frac{p(l | w)}{p(l)}\bigg),
\end{equation} 
where $p(l | w)$ is estimated by $\frac{\operatorname{count}(w, l)}{\operatorname{count}(w)}$, $p(l)$ by $\frac{\operatorname{count}(l)}{|D|}$, $p(w,l)$ by $\frac{\operatorname{count}(w,l)}{|D|}$ and $|D|$ is the number of occurrences of all n-grams in the dataset.

\tabref{tbl:top-20-lmi-refutes} shows that the top LMI-ranked n-grams that are highly correlated with the \textsc{Refutes} class in the training set exhibit a similar correlation in the development set. 
Most of the n-grams express strong negations, which, in hindsight, is not surprising as these idiosyncrasies are induced by the way annotators altered the original claims to generate fake claims.

\paragraph{World Knowledge}
Word embeddings encompass world knowledge, which might augment the performance of claim-only classifiers. To factor out the contribution of world knowledge, we trained two versions of claim-only InferSent \cite{poliak2018hypothesis} on the FEVER claims: one with GloVe embeddings \cite{pennington2014glove} and the other with random embeddings.\footnote{We use InferSent because BERT, being pretrained on Wikipedia, comprises world knowledge~\cite{talmor-etal-2019-commonsenseqa}.} 
The performance with random embeddings was 54.1\%, compared to 57.3\% with GloVe, which is still far above the majority baseline (33.3\%). We conjecture that world knowledge is not the main reason for the success of the claim-only classifier.

\section{Towards Unbiased Evaluation}
\label{sec:sym_data}

Based on the analysis above, we conclude that an unbiased verification dataset should exclude `give-away' phrases in one of its inputs and also not allow the system to solely rely on world knowledge.
The dataset should enforce models to validate the claim with respect to the retrieved evidence. 
Particularly, the truth of some claims might change as the evidence varies over time.

For example, the claim \textit{``Halep failed to ever win a Wimbledon title''} was correct until July 19. A fact-checking system that retrieves information from Halep's Wikipedia page should modify its answer to ``false'' after the update that includes information about her 2019 win.


Towards this goal, we create a \textsc{Symmetric Test Set}. For an original claim-evidence pair, we manually generate a synthetic pair that holds the same relation (i.e.\ \textsc{Supports} or \textsc{Refutes}) while expressing a fact that contradicts the original sentences. Combining the \textsc{Original} and \textsc{Generated} pairs, we obtain two new cross pairs that hold the inverse relations (see \figref{fig:intro}). Examples of generated sentences are provided in \tabref{tab:symm_examples}.


This new test set completely eliminates the ability of models to rely on cues from claims. Considering the two labels of this test set\footnote{\textsc{Not Enough Info} cases are easy to generate so we focus on the two other labels.}, the probability of a label given the existence of any n-gram in the claim or in the evidence is $p(l | w)=0.5$, by construction. 

Also, as the example in \figref{fig:intro} demonstrates, in order to perform well on this dataset, a fact verification classifier may still take advantage of world knowledge (e.g.\ geographical locations), but reasoning should only be with respect to the context.

\begin{table*}[t]
\centering
\resizebox{2.08\columnwidth}{!}{%
\begin{tabular}{c|l|l|c}
\textbf{Source} & \multicolumn{1}{c|}{\textbf{Claim}} & \multicolumn{1}{c|}{\textbf{Evidence}} & \textbf{Label} \\ \midrule
\textbf{\textsc{Original}} & \begin{tabular}[c]{@{}l@{}}Tim Roth is an English actor.\\ \end{tabular} & \begin{tabular}[c]{@{}l@{}}Timothy Simon Roth (born 14 May 1961) is\\ an English actor and director.\end{tabular} & \textsc{Supports} \\ \hline
\textbf{\textsc{Generated}} & \begin{tabular}[c]{@{}l@{}}Tim Roth is an American actor.\end{tabular} & \begin{tabular}[c]{@{}l@{}}Timothy Simon Roth (born 14 May 1961) is \\an American actor and director.\end{tabular} & \textsc{Supports} \\ \midrule
\textbf{\textsc{Original}} & \begin{tabular}[c]{@{}l@{}}Aristotle spent time in Athens.\end{tabular} & \begin{tabular}[c]{@{}l@{}}At seventeen or eighteen years of age,\\ he joined Plato's Academy in Athens and\\ remained there until the age of thirty-seven (c. 347 BC).\end{tabular} & \textsc{Supports} \\ \hline
\textbf{\textsc{Generated}}& \begin{tabular}[c]{@{}l@{}}Aristotle did not visit Athens.\end{tabular} & \begin{tabular}[c]{@{}l@{}}At seventeen or eighteen years of age, \\ he missed the opportunity to join Plato's Academy\\ in Athens and never visited the place.\end{tabular} & \textsc{Supports} \\ \midrule 
\textbf{\textsc{Original}} & \begin{tabular}[c]{@{}l@{}}Telemundo is a English-language \\television network.\end{tabular} & \begin{tabular}[c]{@{}l@{}}Telemundo (teleˈmundo)  is an American\\ Spanish-language terrestrial television network owned\\ by Comcast through the NBCUniversal division \\NBCUniversal Telemundo Enterprises.\end{tabular} & \textsc{Refutes} \\ \hline
\textbf{\textsc{Generated}} & \begin{tabular}[c]{@{}l@{}}Telemundo is a Spanish-language\\ television network.\end{tabular} & \begin{tabular}[c]{@{}l@{}}Telemundo (teleˈmundo) is an American\\ English-language terrestrial television network owned\\ by Comcast through the NBCUniversal division \\NBCUniversal Telemundo Enterprises.\end{tabular} & \textsc{Refutes} \\ \midrule
\textbf{\textsc{Original}} & \begin{tabular}[c]{@{}l@{}}Magic Johnson did not \\play for the Lakers.\end{tabular} & \begin{tabular}[c]{@{}l@{}}He played point guard for the Lakers for 13 seasons.\end{tabular} & \textsc{Refutes} \\ \hline
\textbf{\textsc{Generated}} &\begin{tabular}[c]{@{}l@{}} Magic Johnson played \\for the Lakers.\end{tabular} & He played for the Giants and no other team. & \textsc{Refutes} \\ \bottomrule
\end{tabular}
}
\caption{Examples of pairs from the Symmetric Dataset. Each generated claim-evidence pair holds the relation described in the right column. Crossing the generated sentences with the original ones creates two additional cases with an opposite label (see \figref{fig:intro}).}
\label{tab:symm_examples}
\end{table*}

\section{Towards Unbiased Training}
\label{sec:unbias}

Creating a large symmetric dataset for training is outside the scope of this paper as it would be too expensive. Instead, we propose an algorithmic solution to alleviate the bias introduced by `give-away' n-grams present in the claims. We re-weight the instances in the dataset to flatten the correlation of claim n-grams with respect to the labels.
Specifically, for `give-away' phrases of a particular label, we increase the importance of claims with different labels containing those phrases.

We assign an additional (positive) balancing weight $\alpha^{(i)}$ to each training example $\{x^{(i)},y^{(i)}\}$, determined by the words in the claim.

\paragraph{Bias in the Re-Weighted Dataset} For each n-gram $w_{j}$ in the vocabulary $V$ of the claims, we define the bias towards class $c$ to be of the form:
\begin{equation}
b_{j}^{c} = \frac{\sum_{i=1}^{n} I_{[w_{j}^{(i)}]}(1+\alpha^{(i)})I_{[y^{(i)}=c]}}{\sum_{i=1}^{n} I_{[w_{j}^{(i)}]}(1+\alpha^{(i)})},
\end{equation}
where $I_{[w_{j}^{(i)}]}$ and $I_{[y^{(i)}=c]}$ are the indicators for $w_{j}$ being present in the claim from $x^{(i)}$ and label $y^{(i)}$ being of class $c$, respectively.

\paragraph{Optimization of the Overall Bias}
Finding the $\alpha$ values which minimize the bias leads us to solving the following objective:
\begin{equation} \label{eq:bias_obj}
\min\left(\sum\limits_{j=1}^{\mid V\mid}  \max\limits_{c} (b_{j}^{c}) + \lambda  \|\vec \alpha \|_{2} \right).
\end{equation} 

\paragraph{Re-Weighted Training Objective} 
We calculate the $\alpha$ values separately from the model optimization, as a pre-processing step, by optimizing \eqref{eq:bias_obj}. Using these values, the training objective is re-weighted from the standard $\sum_{i=1}^{n}L(x^{(i)},y^{(i)})$ to
\begin{equation} \label{eq:weight_loss}
  \sum_{i=1}^{n}(1+\alpha^{(i)})L(x^{(i)},y^{(i)}).
\end{equation} 

This re-weighting is independent of the model architecture and can be easily added to any objective, similar to \citet{jiang2019identifying} where they learn instance weights to address labeling bias in datasets.

\section{Experiments}
\label{sec:experimental_setup}

We use the \textsc{Symmetric Test Set} to (1) investigate whether top performing sequence classification models trained on the FEVER dataset are actually verifying claims in the context of evidence; and (2) measure the impact of the re-weighting method described in \cref{sec:unbias} over a classifier.

To achieve the first goal, we use three classifiers. The first is a pre-trained, current FEVER state-of-the-art classifier, \textbf{NSMN}~\citep{nie2018combining} which is a variation of the ESIM~\cite{chen2017enhanced} model, with a number of additional features, such as contextual word embeddings~\citep{peters2018deep}. In addition, we train our own \textbf{ESIM} model with GloVe embeddings, using the available code from~\citet{Gardner2017AllenNLP}. The third is a \textbf{BERT} classifier\footnote{\url{https://github.com/huggingface/pytorch-pretrained-BERT}} that we fine-tune for 3 epochs to classify the relation based on the concatenation of the claim and evidence (with a delimiter token). To measure the impact of our regularization method, we also train the ESIM and BERT models with the re-weighting method.

\paragraph{Symmetric Test Set} The full \textsc{Symmetric Test Set} consists of 956 claim-evidence pairs, created following the procedure described in \cref{sec:sym_data}. The new pairs originated from 99 \textsc{Supports} and 140 \textsc{Refutes} pairs that were randomly picked from the cases which NSMN correctly predicts.\footnote{Due to our focus on the performance drop with respect to the newly generated pairs rather than on the intention of multiplying the difficulties for the top performing model.} After its generation, we asked two subjects to annotate randomly sampled 285 claim-evidence pairs (i.e.\ 30\% of the total pairs in \textsc{Symmetric Test Set}) with one label among \textsc{Supports}, \textsc{Refutes} or \textsc{Not Enough Info}, flagging non-grammatical cases. They agreed with the dataset labels in 94\% of cases, attaining a Cohen $\kappa$ of 0.88 \citep{cohen1960coefficient}. Typos and small grammatical errors were reported in 2\% of the cases. Given the small size of this dataset, we only use it as a test set.

\paragraph{Results}
\label{sec:results}

\begin{table}[t]
\centering
\begin{tabular}{lcc|cc}

 \toprule
 & \multicolumn{2}{c|}{FEVER \textsc{Dev}}& \multicolumn{2}{c}{\textsc{Generated}} \\

Model & \textsc{base}& \textsc{r.w} & \textsc{base} & \textsc{r.w} \\
\midrule

NSMN & 81.8 & - & 58.7 & - \\
ESIM & 80.8 & 76.0 & 55.9 & 59.3 \\
BERT & \textbf{86.2} & 84.6 & 58.3 & \textbf{61.6} \\
\bottomrule
\end{tabular}

\caption{Classifiers' accuracy on the \textsc{Supports} and \textsc{Refutes} cases from the FEVER \textsc{Dev} set and on the \textsc{Generated} pairs for the \textsc{Symmetric Test Set} in the setting of without (\textsc{base}) and with (\textsc{r.w}) re-weight.}
\label{tab:sym_res}
\end{table}

\tabref{tab:sym_res} summarizes the performance of the three models on the \textsc{Supports} and \textsc{Refutes} pairs from the FEVER \textsc{Dev} set and on the created \textsc{Symmetric Test Set} pairs. 
All models perform relatively well on FEVER \textsc{Dev} but achieve less than 60\% accuracy on the synthetic ones. We conjecture that the drop in performance is due to training data bias that is also observed in the development set (see \cref{sec:bias_analysis}) but not in the generated symmetric cases.

Our re-weighting method (\cref{sec:unbias}) helps to reduce the bias in the claims. In \tabref{tbl:reweight_bigrams}, we revisit the give-away bigrams from  \tabref{tbl:top-20-lmi-refutes}. Applying the weights obtained by optimizing  \eqref{eq:bias_obj}, the weighted distribution of these phrases being associated with a specific label in the training set is now roughly uniform.

The re-weighting method increases the accuracy of the ESIM and BERT models by an absolute 3.4\% and 3.3\% respectively. One can notice that this improvement comes at a cost in the accuracy over the FEVER \textsc{Dev} pairs. Again, this can be explained by the bias in the training data that translates to the development set, allowing FEVER-trained models to leverage it. Applying the regularization method, using the same training data, helps to train a more robust model that performs better on our test set, where verification in context is a key requirement.

\section{Related Work}

Large scale datasets are fraught with give-away phrases~\cite{mccoy2019right, niven-kao-2019-probing}. Crowd workers tend to adopt heuristics when creating examples, introducing bias in the dataset. In SNLI (Stanford Natural Language Inference) \cite{bowman2015large}, entailment based solely on the hypothesis forms a very strong baseline ~\cite{poliak2018hypothesis,gururangan2018annotation}. 

Similarly, as shown by \citet{kaushik2018much}, reading comprehension models that rely only on the question (or only on the passage referred to by the question) perform exceedingly well on several popular datasets~\cite{weston2015towards,onishi2016did,hill2015goldilocks}.
To address deficiencies in the SQuAD dataset~\citep{jia2017adversarial}, researchers have proposed approaches for augmenting the existing dataset~\citep{rajpurkar2018know}. In most cases, these augmentations are done manually, and involve constructing challenging examples for existing systems.

\section{Conclusion}

\begin{table}[t!]
\centering
\resizebox{\columnwidth}{!}{%
\begin{tabular}{lcc}
\toprule
\textbf{Bigram}    & \textsc{r.w} \textbf{LMI$ \cdot 10^{-6}$}   & \textsc{r.w} \textbf{$p(l |w)$}  \\ 
\midrule
did not            & 144           & 0.35      \\
yet to             & 30            & 0.33      \\
does not           & 67            & 0.35      \\
refused to         & 55            & 0.35      \\
failed to          & 31            & 0.33      \\
only ever          & 9            & 0.31      \\
incapable being    & 32            & 0.33      \\
to be              & 8            & 0.30      \\
unable to          & 10            & 0.32      \\
not have           & 41            & 0.35      \\

\bottomrule             
\end{tabular}%
}
\caption{Re-weighted statistics ($l=\textsc{Refutes}$) for the bigrams from \tabref{tbl:top-20-lmi-refutes}.  The weights were obtained following the optimization of \eqref{eq:bias_obj} on the training set which contains three labels.}
\label{tbl:reweight_bigrams}
\end{table}

This paper demonstrates that the FEVER dataset contains idiosyncrasies that can be easily exploited by fact-checking classifiers to obtain high classification accuracies. 
Evaluating the claim-evidence reasoning of these models necessitates unbiased datasets.
Therefore, we suggest a way to turn the evaluation FEVER pairs into symmetric combinations for which a decision that is solely based on the claim is equivalent to a random guess. Tested on these pairs, FEVER-trained models show degraded performance. 
To address this problem, we propose a simple method that supports a more robust generalization in the presence of bias.

Moving forward, we suggest using our symmetric dataset in addition to the current retrieval-based FEVER evaluation pipeline. This way, models could be tested both for their evidence retrieval and classification accuracy and for performing the reasoning with respect to the evidence.


\section{Acknowledgments}
We thank the MIT NLP group and the reviewers for their helpful discussion and comments.
This work is supported by DSO grant DSOCL18002.

\bibliography{emnlp-ijcnlp-2019}

\begin{thebibliography}{23}
\expandafter\ifx\csname natexlab\endcsname\relax\def\natexlab#1{#1}\fi

\bibitem[{Alhindi et~al.(2018)Alhindi, Petridis, and
  Muresan}]{alhindi-etal-2018-evidence}
Tariq Alhindi, Savvas Petridis, and Smaranda Muresan. 2018.
\newblock \href {https://doi.org/10.18653/v1/W18-5513} {Where is your evidence:
  Improving fact-checking by justification modeling}.
\newblock In \emph{Proceedings of the First Workshop on Fact Extraction and
  {VER}ification ({FEVER})}, pages 85--90, Brussels, Belgium. Association for
  Computational Linguistics.

\bibitem[{Bowman et~al.(2015)Bowman, Angeli, Potts, and
  Manning}]{bowman2015large}
Samuel~R. Bowman, Gabor Angeli, Christopher Potts, and Christopher~D. Manning.
  2015.
\newblock \href {https://doi.org/10.18653/v1/D15-1075} {A large annotated
  corpus for learning natural language inference}.
\newblock In \emph{Proceedings of the 2015 Conference on Empirical Methods in
  Natural Language Processing}, pages 632--642. Association for Computational
  Linguistics.

\bibitem[{Chen et~al.(2017)Chen, Zhu, Ling, Wei, Jiang, and
  Inkpen}]{chen2017enhanced}
Qian Chen, Xiaodan Zhu, Zhen-Hua Ling, Si~Wei, Hui Jiang, and Diana Inkpen.
  2017.
\newblock \href {http://www.aclweb.org/anthology/P17-1152} {Enhanced lstm for
  natural language inference}.
\newblock In \emph{Proceedings of the 55th Annual Meeting of the Association
  for Computational Linguistics (Volume 1: Long Papers)}, pages 1657--1668.

\bibitem[{Cohen(1960)}]{cohen1960coefficient}
Jacob Cohen. 1960.
\newblock A coefficient of agreement for nominal scales.
\newblock \emph{Educational and psychological measurement}, 20(1):37--46.

\bibitem[{Devlin et~al.(2019)Devlin, Chang, Lee, and
  Toutanova}]{devlin2018bert}
Jacob Devlin, Ming-Wei Chang, Kenton Lee, and Kristina Toutanova. 2019.
\newblock \href {https://doi.org/10.18653/v1/N19-1423} {{BERT}: Pre-training of
  deep bidirectional transformers for language understanding}.
\newblock In \emph{Proceedings of the 2019 Conference of the North {A}merican
  Chapter of the Association for Computational Linguistics: Human Language
  Technologies, Volume 1 (Long and Short Papers)}, pages 4171--4186,
  Minneapolis, Minnesota. Association for Computational Linguistics.

\bibitem[{Evert(2005)}]{evert2005statistics}
Stefan Evert. 2005.
\newblock The statistics of word cooccurrences: word pairs and collocations.

\bibitem[{Gardner et~al.(2017)Gardner, Grus, Neumann, Tafjord, Dasigi, Liu,
  Peters, Schmitz, and Zettlemoyer}]{Gardner2017AllenNLP}
Matt Gardner, Joel Grus, Mark Neumann, Oyvind Tafjord, Pradeep Dasigi,
  Nelson~F. Liu, Matthew Peters, Michael Schmitz, and Luke~S. Zettlemoyer.
  2017.
\newblock \href {http://arxiv.org/abs/arXiv:1803.07640} {Allennlp: A deep
  semantic natural language processing platform}.

\bibitem[{Gururangan et~al.(2018)Gururangan, Swayamdipta, Levy, Schwartz,
  Bowman, and Smith}]{gururangan2018annotation}
Suchin Gururangan, Swabha Swayamdipta, Omer Levy, Roy Schwartz, Samuel Bowman,
  and Noah~A. Smith. 2018.
\newblock \href {https://doi.org/10.18653/v1/N18-2017} {Annotation artifacts in
  natural language inference data}.
\newblock In \emph{Proceedings of the 2018 Conference of the North American
  Chapter of the Association for Computational Linguistics: Human Language
  Technologies, Volume 2 (Short Papers)}, pages 107--112. Association for
  Computational Linguistics.

\bibitem[{Hill et~al.(2016)Hill, Bordes, Chopra, and
  Weston}]{hill2015goldilocks}
Felix Hill, Antoine Bordes, Sumit Chopra, and Jason Weston. 2016.
\newblock \href {https://arxiv.org/abs/1511.02301} {The goldilocks principle:
  Reading children's books with explicit memory representations}.
\newblock In \emph{International Conference on Learning Representations 2016}.

\bibitem[{Jia and Liang(2017)}]{jia2017adversarial}
Robin Jia and Percy Liang. 2017.
\newblock \href {https://doi.org/10.18653/v1/D17-1215} {Adversarial examples
  for evaluating reading comprehension systems}.
\newblock In \emph{Proceedings of the 2017 Conference on Empirical Methods in
  Natural Language Processing}, pages 2021--2031. Association for Computational
  Linguistics.

\bibitem[{Jiang and Nachum(2019)}]{jiang2019identifying}
Heinrich Jiang and Ofir Nachum. 2019.
\newblock \href {https://arxiv.org/abs/1901.04966} {Identifying and correcting
  label bias in machine learning}.
\newblock \emph{arXiv preprint arXiv:1901.04966}.

\bibitem[{Kaushik and Lipton(2018)}]{kaushik2018much}
Divyansh Kaushik and Zachary~C. Lipton. 2018.
\newblock \href {http://aclweb.org/anthology/D18-1546} {How much reading does
  reading comprehension require? a critical investigation of popular
  benchmarks}.
\newblock In \emph{Proceedings of the 2018 Conference on Empirical Methods in
  Natural Language Processing}, pages 5010--5015. Association for Computational
  Linguistics.

\bibitem[{McCoy et~al.(2019)McCoy, Pavlick, and Linzen}]{mccoy2019right}
Tom McCoy, Ellie Pavlick, and Tal Linzen. 2019.
\newblock \href {https://www.aclweb.org/anthology/P19-1334} {Right for the
  wrong reasons: Diagnosing syntactic heuristics in natural language
  inference}.
\newblock In \emph{Proceedings of the 57th Annual Meeting of the Association
  for Computational Linguistics}, pages 3428--3448, Florence, Italy.
  Association for Computational Linguistics.

\bibitem[{Nie et~al.(2019)Nie, Chen, and Bansal}]{nie2018combining}
Yixin Nie, Haonan Chen, and Mohit Bansal. 2019.
\newblock \href {https://arxiv.org/abs/1811.07039} {Combining fact extraction
  and verification with neural semantic matching networks}.
\newblock In \emph{Association for the Advancement of Artificial Intelligence}.

\bibitem[{Niven and Kao(2019)}]{niven-kao-2019-probing}
Timothy Niven and Hung-Yu Kao. 2019.
\newblock \href {https://www.aclweb.org/anthology/P19-1459} {Probing neural
  network comprehension of natural language arguments}.
\newblock In \emph{Proceedings of the 57th Annual Meeting of the Association
  for Computational Linguistics}, pages 4658--4664, Florence, Italy.
  Association for Computational Linguistics.

\bibitem[{Onishi et~al.(2016)Onishi, Wang, Bansal, Gimpel, and
  McAllester}]{onishi2016did}
Takeshi Onishi, Hai Wang, Mohit Bansal, Kevin Gimpel, and David~A. McAllester.
  2016.
\newblock \href {https://aclweb.org/anthology/D16-1241} {Who did what: {A}
  large-scale person-centered cloze dataset}.
\newblock In \emph{{Proceedings of the 2016 Conference on Empirical Methods in
  Natural Language Processing}}, pages 2230--2235. The Association for
  Computational Linguistics.

\bibitem[{Pennington et~al.(2014)Pennington, Socher, and
  Manning}]{pennington2014glove}
Jeffrey Pennington, Richard Socher, and Christopher Manning. 2014.
\newblock \href {http://www.aclweb.org/anthology/D14-1162} {Glove: Global
  vectors for word representation}.
\newblock In \emph{Proceedings of the 2014 conference on empirical methods in
  natural language processing}, pages 1532--1543.

\bibitem[{Peters et~al.(2018)Peters, Neumann, Iyyer, Gardner, Clark, Lee, and
  Zettlemoyer}]{peters2018deep}
Matthew Peters, Mark Neumann, Mohit Iyyer, Matt Gardner, Christopher Clark,
  Kenton Lee, and Luke Zettlemoyer. 2018.
\newblock \href {http://www.aclweb.org/anthology/N18-1202} {Deep contextualized
  word representations}.
\newblock In \emph{Proceedings of the 2018 Conference of the North American
  Chapter of the Association for Computational Linguistics: Human Language
  Technologies, Volume 1 (Long Papers)}, pages 2227--2237.

\bibitem[{Poliak et~al.(2018)Poliak, Naradowsky, Haldar, Rudinger, and
  Van~Durme}]{poliak2018hypothesis}
Adam Poliak, Jason Naradowsky, Aparajita Haldar, Rachel Rudinger, and Benjamin
  Van~Durme. 2018.
\newblock \href {https://doi.org/10.18653/v1/S18-2023} {Hypothesis only
  baselines in natural language inference}.
\newblock In \emph{Proceedings of the Seventh Joint Conference on Lexical and
  Computational Semantics}, pages 180--191, New Orleans, Louisiana. Association
  for Computational Linguistics.

\bibitem[{Rajpurkar et~al.(2018)Rajpurkar, Jia, and Liang}]{rajpurkar2018know}
Pranav Rajpurkar, Robin Jia, and Percy Liang. 2018.
\newblock \href {http://aclweb.org/anthology/P18-2124} {Know what you don't
  know: Unanswerable questions for squad}.
\newblock In \emph{Proceedings of the 56th Annual Meeting of the Association
  for Computational Linguistics (Volume 2: Short Papers)}, pages 784--789.
  Association for Computational Linguistics.

\bibitem[{Talmor et~al.(2019)Talmor, Herzig, Lourie, and
  Berant}]{talmor-etal-2019-commonsenseqa}
Alon Talmor, Jonathan Herzig, Nicholas Lourie, and Jonathan Berant. 2019.
\newblock \href {https://doi.org/10.18653/v1/N19-1421} {{C}ommonsense{QA}: A
  question answering challenge targeting commonsense knowledge}.
\newblock In \emph{Proceedings of the 2019 Conference of the North {A}merican
  Chapter of the Association for Computational Linguistics: Human Language
  Technologies, Volume 1 (Long and Short Papers)}, pages 4149--4158,
  Minneapolis, Minnesota. Association for Computational Linguistics.

\bibitem[{Thorne et~al.(2018)Thorne, Vlachos, Christodoulopoulos, and
  Mittal}]{thorne2018fever}
James Thorne, Andreas Vlachos, Christos Christodoulopoulos, and Arpit Mittal.
  2018.
\newblock \href {http://www.aclweb.org/anthology/N18-1074} {Fever: a
  large-scale dataset for fact extraction and verification}.
\newblock In \emph{Proceedings of the 2018 Conference of the North American
  Chapter of the Association for Computational Linguistics: Human Language
  Technologies, Volume 1 (Long Papers)}, pages 809--819.

\bibitem[{Weston et~al.(2016)Weston, Bordes, Chopra, Rush, van Merri{\"e}nboer,
  Joulin, and Mikolov}]{weston2015towards}
Jason Weston, Antoine Bordes, Sumit Chopra, Alexander~M Rush, Bart van
  Merri{\"e}nboer, Armand Joulin, and Tomas Mikolov. 2016.
\newblock \href {https://arxiv.org/abs/1502.05698} {Towards ai-complete
  question answering: A set of prerequisite toy tasks}.
\newblock In \emph{International Conference on Learning Representations}.

\end{thebibliography}
\bibliographystyle{acl_natbib}

\clearpage
\appendix
\section{Claim-only vs. Evidence-aware Classification}

\tabref{tab:claim_only_res} shows the performance of the claim-only BERT classifier and numerous evidence-aware baseline in a three-class (\textsc{Support}, \textsc{Refute}, \textsc{Not Enough Info}) setting. 

\begin{table}[H]
\centering
\small
\begin{tabular}{l|c}
\textbf{Model}     & \textbf{Accuracy}          \\ \hline
Majority Baseline                      & 33.3              \\ \hline \hline
\multicolumn{2}{c}{\textbf{Evidence-Aware Classifiers}} \\ \hline
DA             & 52.1              \\ 
NSMN                           & 69.7              \\ \hline \hline
\multicolumn{2}{c}{\textbf{Claim-Only Aware Classifiers}}      \\ \hline
InferSent (random emb.)                  & 54.1              \\ 
InferSent (GloVe)                       & 57.3              \\ 
BERT               & 61.7              \\ \hline
\end{tabular}
\caption{Results of evidence-aware and claim-only classifiers on the three label development set of the FEVER dataset.} \label{tab:claim_only_res}
\end{table}

\section{Additional Analysis}
\label{appx:appendix_one}

\subsection{Fever Split}
\label{appx:fever_split}

The split of the public FEVER dataset is described in \tabref{tbl:fever_split}.

\begin{table}[H]
\centering
\resizebox{\columnwidth}{!}{%
\begin{tabular}{lccc}
\\ 
\textbf{Split}    & \textbf{\textsc{Support}}   & \textbf{\textsc{Refute}}  & \textbf{\textsc{Not Enough Info}}    \\ \hline
Training            & 80,035           & 29,775              & 35,639       \\
Development            & 6,666           & 6,666              & 6,666       \\ \hline
Total            & 86,701           & 36,441              & 42,305 \\ \hline
\end{tabular}%
}
\caption{Fever dataset split.}
\label{tbl:fever_split}
\end{table}

\subsection{Top LMI-ranked Bigrams in Train and Development Set}
\label{appx:top_bigrams}

\tabref{tbl:top-20-LMI-supports} and \tabref{tbl:top-20-LMI-nei} summarize the top 10 bigrams for \textsc{Support} and \textsc{Not Enough Info}. The correlation between the biased phrases in the two dataset splits is not as strong as in the \textsc{Refute} label, presented in the paper. However, one can notice that some of the biased bigrams in the training set, such as ``least one'' and ``starred movie'', translate to cues that can help in predictions over the development set. Bigrams are chosen for this exploratory analysis as they yield more comprehensible phrases.

\begin{table}[H]
\centering
\resizebox{\columnwidth}{!}{
\begin{tabular}{lcc|cc}
\textbf{}         & \multicolumn{2}{c|}{\textbf{Train}} & \multicolumn{2}{c}{\textbf{Development}} \\ \cline{2-5}
\textbf{Bigram}   & \textbf{LMI} $\cdot10^{-6}$  & \textbf{$p(l | w)$}  & \textbf{LMI} $\cdot10^{-6}$    & \textbf{$p(l |w)$}    \\ \hline
united states     & 271           & 0.64    & 268              & 0.44       \\
least one         & 269           & 0.90    & 267              & 0.77       \\
at least          & 256           & 0.72    & 163              & 0.48       \\
person who        & 162           & 0.90    & 135              & 0.61       \\
stars actor       & 143           & 0.86    & 111              & 0.71       \\
won award         & 133           & 0.80    & 50               & 0.56       \\
american actor    & 126           & 0.79    & 55               & 0.45       \\
starred movie     & 100           & 0.88    & 34               & 0.80       \\
from united       & 100           & 0.82    & 108              & 0.67       \\
from america      & 96            & 0.89    & 108              & 0.74      \\
\hline
\end{tabular}
}
\caption{Top 10 LMI-ranked bigrams in the train set of FEVER for \textsc{Support}.}
\label{tbl:top-20-LMI-supports}
\end{table}

\begin{table}[H]
\centering
\resizebox{\columnwidth}{!}{%
\begin{tabular}{lcc|cc}
\textbf{}        & \multicolumn{2}{c|}{\textbf{Train}} & \multicolumn{2}{c}{\textbf{Development}} \\ \cline{2-5}
\textbf{Bigram}   & \textbf{LMI} $\cdot10^{-6}$  & \textbf{$p(l | w)$}  & \textbf{LMI} $\cdot10^{-6}$    & \textbf{$p(l |w)$}  \\ \hline
worked with      & 221           & 0.40       & 129              & 0.56   \\
s name           & 99            & 0.59       & 106              & 0.65   \\
award winning    & 98            & 0.52       & 208              & 0.79   \\
wyatt earp       & 96            & 0.42       & *                & 0.00   \\
finished college & 86            & 0.68       & 10               & 0.42   \\
and it           & 86            & 0.42       & 254              & 0.73   \\
will ferrell     & 79            & 0.46       & *                & 0.00   \\
can be           & 75            & 0.35       & 72               & 0.48   \\
and he           & 74            & 0.38       & 52               & 0.59   \\
tim rice         & 70            & 0.41       & *                & 0.00   \\
\hline
\end{tabular}%
}
\caption{Top 10 LMI-ranked bigrams in the train set of FEVER for \textsc{Not Enough Info}. * denotes computationally infeasible, as occurrence is zero in the development set.}
\label{tbl:top-20-LMI-nei}
\end{table}

\begin{figure}[H]
    \centering
 \includegraphics[width=0.47\textwidth]{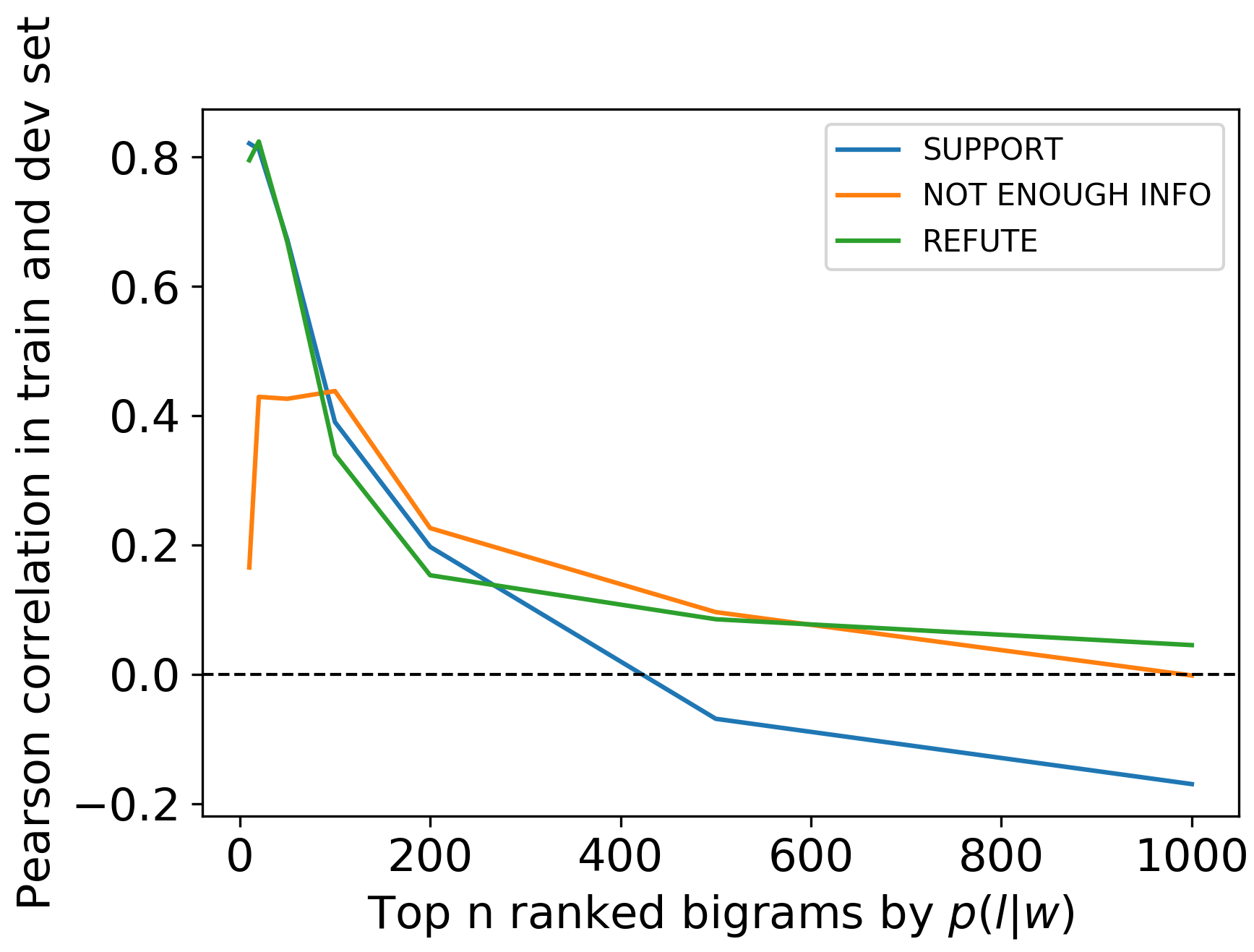}
 \caption{Pearson $r$ scores of $p(l|w)$ for the top LMI-ranked bigrams in the train and development sets. }  \label{fig:cp_corr}
\end{figure}

We calculated the Pearson correlation score between $p(l|w$) for both train and development set. \figref{fig:cp_corr} shows curves that start from very high correlations (i.e. 0.8 to 0.5) for the top $p(l|w)$-ranked $\sim$50-100 bigrams of \textsc{Refute} and \textsc{Support} (the curve for \textsc{Not Enough Info} is less stable), dropping at around rank 400, supporting the existence of `give-away-bigrams' and that they are common in both training and development set.


\subsection{Top Bigram Distribution in the Development Claims}

\begin{figure}[H]
    \centering
  \includegraphics[width=0.48\textwidth]{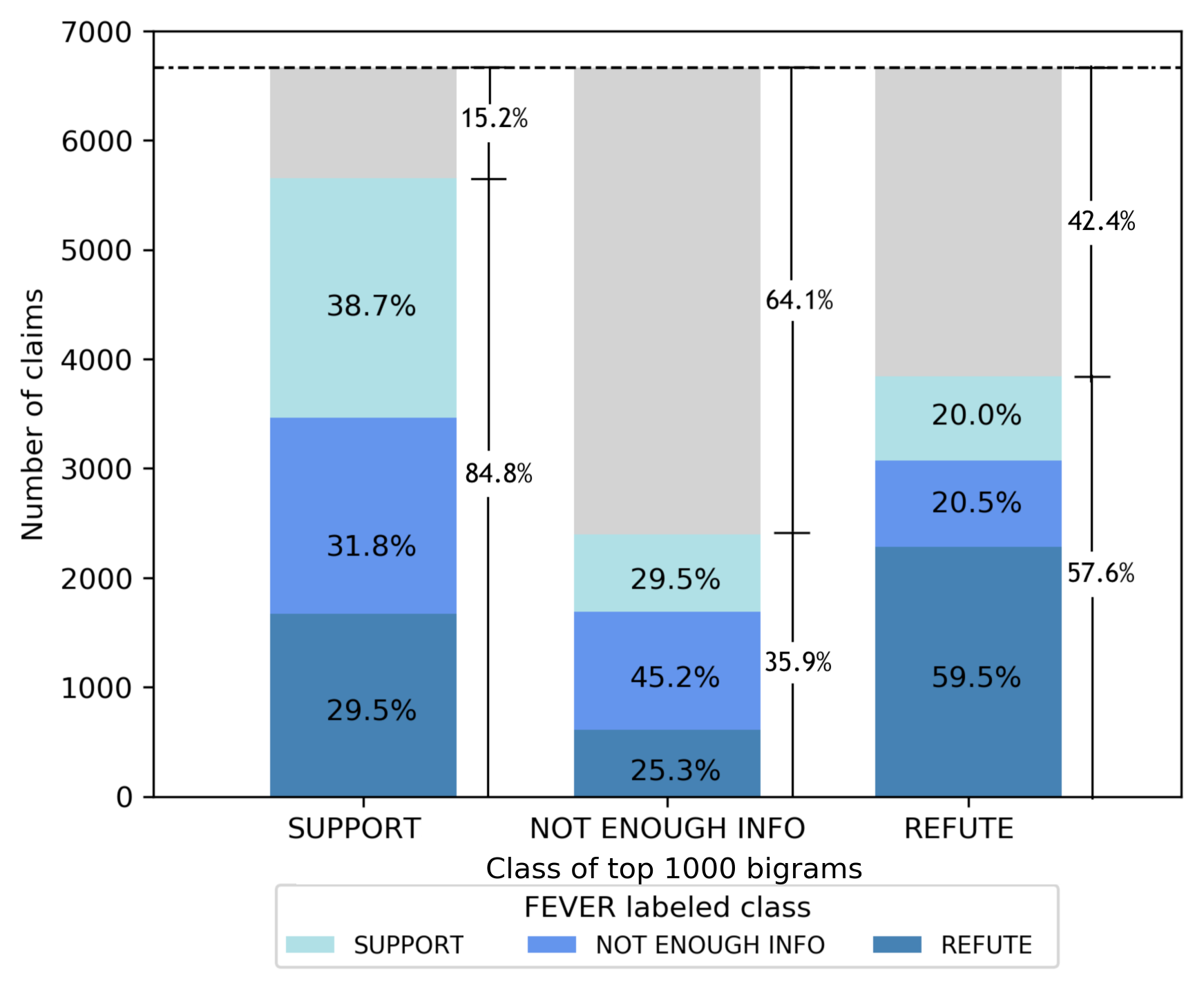}
  \caption{Percentage of claims containing at least one of the top 1,000 LMI-ranked bigrams (colors are used to express the class the claims were associated to). The overall heights of the bars indicate the number of claims expected for each class (i.e. 6,666).} \label{fig:stacked_gold}
\end{figure}

\figref{fig:stacked_gold} illustrates the distribution of the top 1,000 LMI-ranked training set bigrams in the development set. In the case of the \textsc{Refute} class, we see that 57.6\% of the \textsc{Refute} claims in the development set contain the top 1,000 LMI-ranked bigrams. Out of them, a high 59.5\% are indeed labeled \textsc{Refute}. This concludes that 34.3\% of all \textsc{Refute} claims are potentially biased. Following the same line of explanation, 32.8\% and 16.2\% of the \textsc{Support} and \textsc{Not Enough Info} claims also face this problem. 

\end{document}